\documentclass[conference]{IEEEtran}
\IEEEoverridecommandlockouts
\usepackage{cite}
\usepackage{amsmath,amssymb,amsfonts}
\usepackage{algorithmic}
\usepackage{graphicx}
\usepackage{textcomp}
\usepackage{xcolor}
\def\BibTeX{{\rm B\kern-.05em{\sc i\kern-.025em b}\kern-.08em
    T\kern-.1667em\lower.7ex\hbox{E}\kern-.125emX}}
\usepackage{cite}
\usepackage[colorlinks=true, linkcolor=blue, citecolor=blue, urlcolor=blue]{hyperref}
\usepackage{url}
\usepackage{orcidlink}
\usepackage{authblk} 
\usepackage{multirow}
\usepackage{booktabs}    
\usepackage{makecell}   

\begin{document}

\title{Neural-MCRL: Neural Multimodal Contrastive Representation Learning for EEG-based Visual Decoding}


\author{
	Yueyang Li\textsuperscript{1,$\star$},
  Zijian Kang\textsuperscript{1,$\star$},
	Shengyu Gong\textsuperscript{1},
	Wenhao Dong\textsuperscript{1},
    Weiming Zeng\textsuperscript{1,$\dagger$},\\
	Hongjie Yan\textsuperscript{2},
	Wai Ting Siok\textsuperscript{3},
	and Nizhuan Wang\textsuperscript{3,$\dagger$}
    \thanks{Accepted by The 26th IEEE International Conference on Multimedia and Expo (ICME 2025)}
	\thanks{$\star$: Yueyang Li and Zijian Kang are co-first authors. $\dagger$: Weiming Zeng and Nizhuan Wang are corresponding authors. This work was supported by the National Natural Science Foundation of China (grant number 31870979), the Hong Kong Polytechnic University Start-up Fund (Project ID: P0053210), the Hong Kong Polytechnic University Faculty Reserve Fund (Project ID: P0053738), and an internal grant from the Hong Kong Polytechnic University (Project ID: P0048377).}

}

\affil{
	\textsuperscript{1}\textit{Lab of Digital Image and Intelligent Computation, Shanghai Maritime University}, Shanghai 201306, China\\
	\textsuperscript{2}\textit{Affiliated Lianyungang Hospital of Xuzhou Medical University}, Lianyungang 222002, China\\
	\textsuperscript{3}\textit{Department of Chinese and Bilingual Studies, The Hong Kong Polytechnic University}, Hong Kong SAR, China
}
\maketitle
\begin{abstract}
Decoding neural visual representations from electroencephalogram (EEG)-based brain activity is crucial for advancing brain-machine interfaces (BMI) and has transformative potential for neural sensory rehabilitation. While multimodal contrastive representation learning (MCRL) has shown promise in neural decoding, existing methods often overlook semantic consistency and completeness within modalities and lack effective semantic alignment across modalities. This limits their ability to capture the complex representations of visual neural responses. We propose Neural-MCRL, a novel framework that achieves multimodal alignment through semantic bridging and cross-attention mechanisms, while ensuring completeness within modalities and consistency across modalities. Our framework also features the Neural Encoder with Spectral-Temporal Adaptation (NESTA), a EEG encoder that adaptively captures spectral patterns and learns subject-specific transformations. Experimental results demonstrate significant improvements in visual decoding accuracy and model generalization compared to state-of-the-art methods, advancing the field of EEG-based neural visual representation decoding in BMI. Codes will be available at: https://github.com/NZWANG/Neural-MCRL.
\end{abstract}


\section{Introduction}
\label{sec:intro}
Decoding neural representations of visual information from human brain activity is instrumental for understanding cognitive processes and advancing brain-machine interfaces (BMI) \cite{du2023decoding}. This process entails accurately and rapidly identifying objects within complex scenes and mapping neural signals to visual semantics \cite{dicarlo2007untangling}. Electroencephalography (EEG) is extensively used in BMI for its high temporal resolution and portability, making it ideal for real-time neural decoding, although it also faces challenges in low signal-to-noise ratio (SNR) and  spatial resolution \cite{li2024tale}. Decoding visual semantics from EEG now heavily relies on incorporating multimodal data to compensate for the limited neural recordings. This is achieved by using rich visual-semantic embeddings \cite{chen2024cinematic,chen2024bridging} created through a technique known as multimodal contrastive representation learning (MCRL). For instance, models like NICE \cite{song2023decoding} and BraVL \cite{du2023decoding} demonstrate promising outcomes by associating neural features with pre-trained image-text embeddings, leveraging extensive language-vision models \cite{radford2021learning} to connect EEG signals with sophisticated conceptual knowledge. However, current MCRL methods often simplify semantic alignment as a simple pairwise matching issue, focusing primarily on cross-modal connections while neglecting semantic consistency and completeness within each modality. Moreover, without adequately considering the internal semantic and conceptual richness of each modality, existing methods risk developing fragmented mappings that do not fully capture the intricate nature of neural responses \cite{tang2023semantic}. In addition, current methods lack effective mechanisms to bridge the semantic gap between modalities and capture their dynamic interactions. They rely solely on static feature matching for alignment, overlooking the context-specific nature of neural responses to visual stimuli. These limitations can reduce the accuracy of visual decoding, hinder generalization to novel stimuli and obscure subtle conceptual relationships.

\begin{figure*}[hpbt]
	\centering
	\includegraphics[width = 0.93\textwidth]{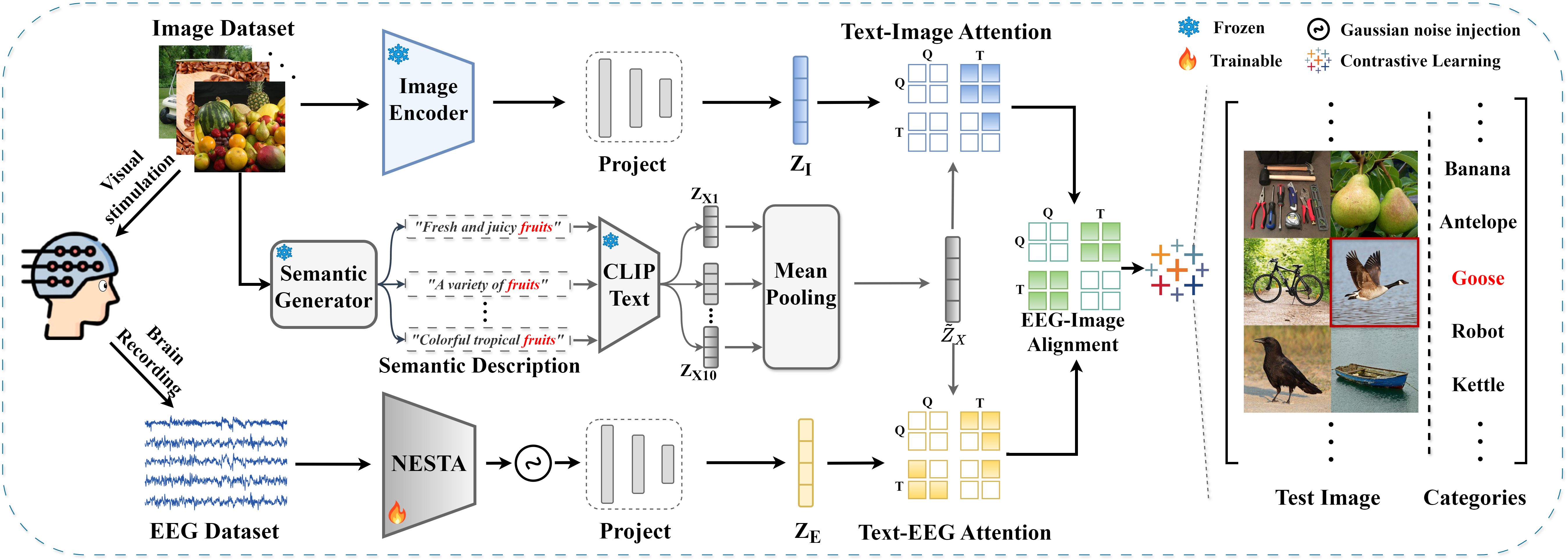}
	\caption{Overall framework of the Neural-MCRL.}
	\label{framework}
\end{figure*}

To address the aforementioned limitations, we propose a novel framework named Neural-MCRL for decoding visual neural representations from EEG signals during object recognition tasks, with a particular emphasis on semantic completeness and robust feature extraction. The main contributions of this work are summarized as follows:
\begin{itemize}
	\item[$\bullet$] The proposed Neural-MCRL is a novel framework that achieves alignment of multimodal contrastive representation by mapping multimodal data into a shared semantic space using semantic bridge and cross-attention mechanisms. It explicitly enhances semantics by ensuring inter-modal semantic consistency and intra-modal semantic completion within the shared embedding space.
	\item[$\bullet$] We propose the Neural Encoder with Spectral-Temporal Adaptation (NESTA), a specialized EEG encoder featuring novel plug-and-play modules that jointly learn subject-specific channel transformations and adaptively capture spectral patterns, preserving critical temporal-spectral information within EEG embeddings.
	\item[$\bullet$] Our approach significantly enhances both decoding accuracy and model generalization in EEG-based visual decoding tasks when compared to current state-of-the-art methods.
\end{itemize}

\section{Related Work}

\subsection{EEG-based Visual Decoding}
Decoding visual representations from neural activity has become a hot research topic to neuroscience and BMI \cite{li2024visual}. Early EEG studies achieved modest recognition performance due to temporal artifacts \cite{ahmed2021object}. In 2022, Gifford et al. constructed a large-scale Visual-EEG dataset with RSVP paradigm to encode complex images into EEG signals, which can be used to distinguish diverse categories from neural activity \cite{gifford2022large}. Further, Du et al. \cite{du2023decoding} proposed a trimodal representation combining visual, linguistic, and neural data for decoding new categories. Song et al. \cite{song2023decoding} developed an EEG encoder with Temporal-Spatial convolution using larger kernels and fewer parameters, to enhance the decoding performance. Additionally, MB2C \cite{wei2024mb2c} employs a dual-GAN architecture to unify modality-related features, while ATM \cite{li2024visual} further aligns EEG with image embeddings to boost performance. Despite these advancements, challenges remain in ensuring cross-subject performance and capturing intrinsic semantic relationships between visual stimuli and neural responses.

\subsection{Multimodal Contrastive Representation Learning}
Advancements in contrastive learning, exemplified by techniques such as SimCLR and MoCo \cite{yeh2022decoupled}, have significantly enhanced representation learning through pair-based alignment. In the multimodal field, CLIP \cite{radford2021learning} demonstrates that aligning language and vision improves joint representations and enables zero-shot knowledge transfer. This concept has been swiftly applied to EEG decoding, where neural signals are aligned with external stimuli for Zero-Shot Neural Decoding \cite{du2023decoding}. Through Zero-Shot learning, EEG signals can be transformed into meaningful representations without task-specific training \cite{mathis2024decoding}, leveraging pre-trained networks to bridge modality gaps \cite{wei2024mb2c, li2024visual}. While reducing modality discrepancies benefits zero-shot classification, the meaningful structural organization in the latent space proves more crucial \cite{jiang2023understanding}.

\section{Method}
\subsection{Problem Formulation}
We consider the dataset $\mathcal{D} = \{ (E_i, I_i, y_i) \}_{i=1}^N$ of each sample consisting of an EEG recording $E_i \in \mathbb{R}^{C \times L}$, visual stimuli $I_i$, and class labels $y_i \in \mathcal{Y}$, where $N$ is the sample count, $C$ is the number of EEG channels, and $L$ is the length of time window. We aim to learn EEG embeddings $Z_E = f(E) \in \mathbb{R}^{N \times F}$ from EEG data $E = \{ E_i \}_{i=1}^N$ using an encoder $f$ that captures subject specificity and EEG dynamics in an $F$-dimensional space. The corresponding visual stimuli $I = \{ I_i \}_{i=1}^N$ and their text descriptions are processed through CLIP to obtain image embeddings $Z_I \in \mathbb{R}^{N \times F}$ and text embeddings $Z_{X} \in \mathbb{R}^{N \times F}$. For zero-shot learning, we partition $\mathcal{D}$ into seen classes $\mathcal{D}_S = \{ 1, \ldots, S \}$ for training and unseen classes $\mathcal{D}_U = \{ S+1, \ldots, S+U \}$ for testing, where $\mathcal{D}_S \cap \mathcal{D}_U = \emptyset$. Training uses only $\mathcal{D}_S$, while inference requires decoding EEG signals in $\mathcal{D}_U$ into images by computing their similarity to novel visual representations in the shared embedding space. As illustrated in Figure \ref{framework}, we present the Neural-MCRL framework to achieve this goal.

\subsection{NESTA for EEG Embedding}
The proposed NESTA aims to encode EEG signals into effective embeddings by incorporating subject-specific transformations and advanced network architectures, as shown in Figure \ref{nest}.

\subsubsection*{\textbf{Subject-specific Layer}}
To capture subject-specific variations and underlying signal characteristics, we introduce a novel subject-specific layer, which uses a learnable, subject-dependent linear transformation across channels. For each subject $s$, the transformation is implemented as: 
\begin{equation}
	E_s = \text{SubjectLayer}(E_i, s) = \mathbf{M}_s E_i, \label{eq1}
\end{equation}
where \( E_s \in \mathbb{R}^{C \times L} \) represents the subject-specific transformed EEG data, and $\mathbf{M}_s \in \mathbb{R}^{C \times C}$ is a learnable weight matrix that captures subject-specific channel interactions. Specifically, the matrices $\mathbf{M}_s$ are initialized as identity matrices, preserving critical patterns in early training while progressively adapting to individual characteristics. 

\subsubsection*{\textbf{Transformer-based Block}}
This block employs a modified iTransformer \cite{liu2023itransformer} to capture temporal and spatial EEG dynamics. This block consequentially consists of Multi-Head Attention Layer via expansion factors and multilayer perceptron (MLP) residual connections, and the Feed Forward Network incorporating \texttt{Conv1d} operations with activation and dropout mechanisms. The block process can be expressed as: 
\begin{equation} 
	E_{\text{t}} = \text{iTransformer}(E_s) \in \mathbb{R}^{B \times C \times L}, \end{equation}
where $B$ is the batch size. The output representation provides a robust foundation for further feature refinement and cross-modal alignment within the NESTA framework.

\subsubsection*{\textbf{Neural-Spectral Adaptation Block}}
\begin{figure}[tbp]
	\centering
	\includegraphics[width = 0.39\textwidth]{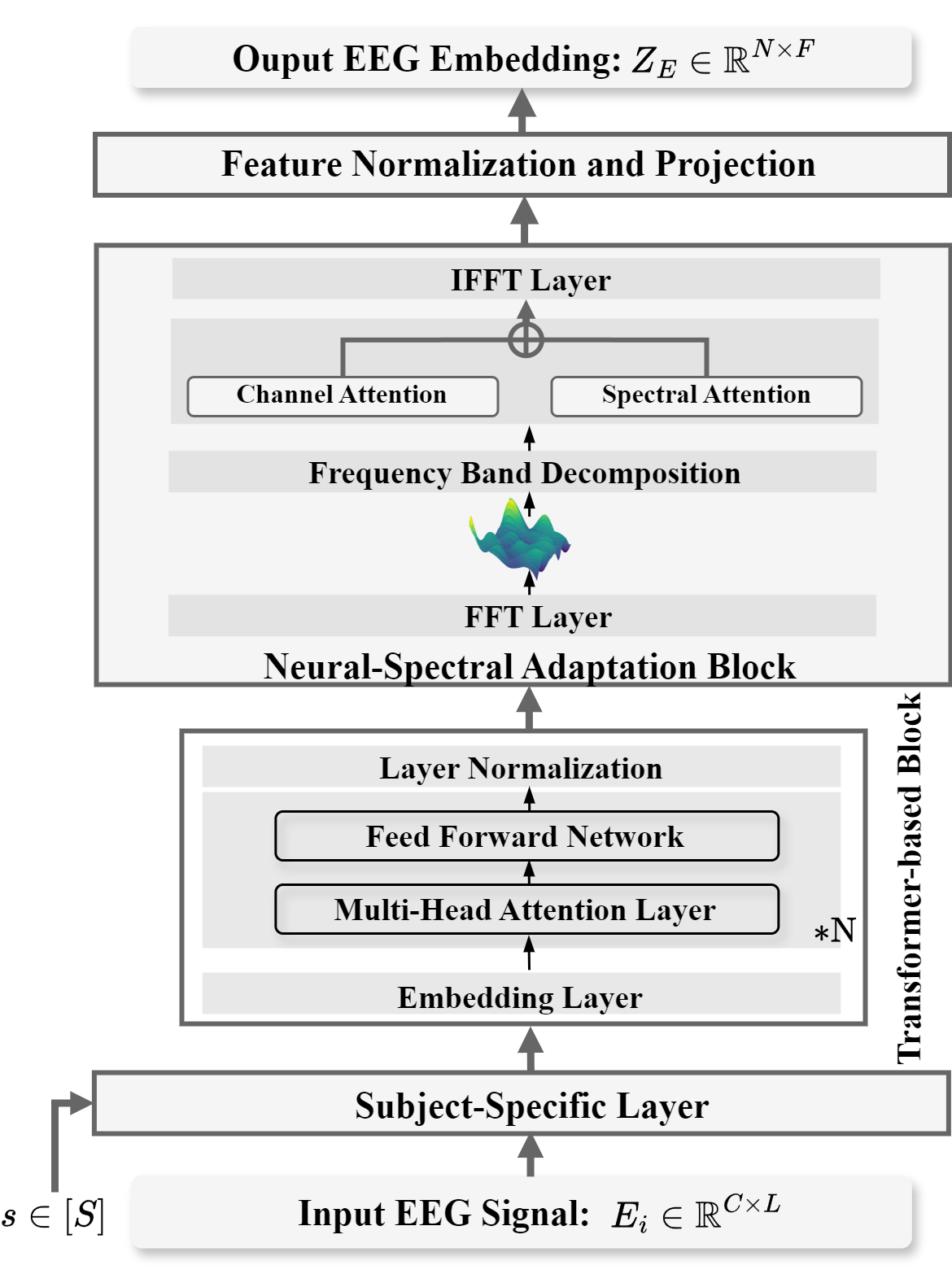}
	\caption{Architecture of the NESTA for EEG embedding.}
	\label{nest}
\end{figure}

We introduced this block that performs adaptive spectral processing with dual attention mechanisms, aiming at learning both channel-wise patterns and frequency-specific features through spectral decomposition and attention mechanisms. Firstly, we transform each channel EEG time series to the frequency domain:
\begin{equation}
	\varGamma = \mathcal{F}[E_{\text{t}}] \in \mathbb{R}^{B \times C \times F},
\end{equation}
where \(\mathcal{F}[\cdot]\) denotes the Fast Fourier Transform (FFT), \(F\) is the transformed sequence length in the frequency domain, and $\varGamma$ encapsulates the frequency spectrum of the EEG time series across all channels. Then, the frequency spectrum corresponding to each channel is decomposed into five canonical EEG bands $\mathcal{B}$: $\delta$, $\theta$, $\alpha$, $\beta$, and $\gamma$. For each band $b \in \mathcal{B}$, we apply frequency masking to compute banded powers as follow:
\begin{align}
	\centering
	\varGamma_{b} &= \varGamma \odot M_b(f), \\
	P_b &= \sum_{f} |\varGamma_b|^2 \in \mathbb{R}^{B \times C},
\end{align}
where $M_b(f)$ is a binary mask for frequency band $b$, and $P_b$ represents the banded power. To capture both spatial and spectral relationships, we employ two attention mechanisms:
\begin{align}
	w_c &= \sigma\left(\mathbf{W}_c\left[\frac{1}{F}\sum_{f=1}^F P_b^{(f)}\right] + \mathbf{b}_c\right) \in \mathbb{R}^{B \times C \times 1}, \\
	w_s &= \text{softmax}\left(\mathbf{W}_s\left[\frac{1}{C}\sum_{c=1}^C P_b^{(c)}\right] + \mathbf{b}_s\right) \in \mathbb{R}^{B \times |\mathcal{B}|},
\end{align}
where $w_c$ denotes channel attention weights and $w_s$ represents spectral attention weights for each frequency band, $\mathbf{W}_c \in \mathbb{R}^{C \times C}$ and $\mathbf{W}_s \in \mathbb{R}^{|\mathcal{B}| \times |\mathcal{B}|}$ are learnable weight matrices with their corresponding bias terms $\mathbf{b}_c \in \mathbb{R}^C$ and $\mathbf{b}_s \in \mathbb{R}^{|\mathcal{B}|}$, $P_b^{(f)}$ and $P_b^{(c)}$ indicate the banded power features aggregated along frequency and channel dimensions, respectively, and $\sigma(\cdot)$ represents the sigmoid activation function. Lastly, the final output integrates the inverse FFT (see IFFT Layer in Figure \ref{nest}) of band-specific features with these attention weights and a learnable residual connection:
\begin{equation}
E_{\text{n}} = \alpha \cdot \text{LN}(\mathcal{F}^{-1}[\sum_{b} X_b \odot w_c \odot w_{s,b}]) + (1-\alpha) \cdot E_{\text{t}},
\end{equation}
where $w_{s,b}$ is the attention weight for frequency band $b$ from $w_s$, and $\alpha$ is a learnable parameter. This design enables the module to adaptively focus on spectral components while preserving temporal information through the residual connection.

\subsubsection*{\textbf{Feature Normalization and Projection}} In this step, the layer normalization is first applied to retrieve the normalized features, which are then projected into a shared embedding space, yielding EEG embeddings $Z_E$ that will align with image and text embeddings to facilitate cross-modal alignment and zero-shot learning.

\subsection{Image Encoder}

To capture visual semantics, we use Vision Transformer as our image encoder, which typically outperforms traditional models in image decoding tasks \cite{grootswagers2022human}. By leveraging the model’s pre-trained weights, we aim to obtain high-dimensional embeddings that robustly represent the underlying information. Subsequently, these embeddings are projected into a shared embedding space, enabling smooth integration with EEG signals.

\subsection{Semantic Enhancement}
\begin{table*}[htbp]
	\centering
	\caption{Overall accuracy (\%) comparison of \textit{N}-way zero-shot classification among competing models: top-1 and top-5.}
	\label{acc}
	\scalebox{0.67}{
		\renewcommand{\arraystretch}{1}
		\begin{tabular}[c]{cccc cccc cccc cccc cccc cccc}
			\hline\noalign{\hrule height 1.2pt}
			& \multicolumn{22}{c}{\normalsize Subject-dependent: train and test on one subject} \\                                                                                                      		  
			\cline{3-24} 
			{Type}                          & {Method}                         & \multicolumn{2}{c}{Subject 1} & \multicolumn{2}{c}{Subject 2} & \multicolumn{2}{c}{Subject 3} & \multicolumn{2}{c}{Subject 4} & \multicolumn{2}{c}{Subject 5} & \multicolumn{2}{c}{Subject 6} & \multicolumn{2}{c}{Subject 7} & \multicolumn{2}{c}{Subject 8} & \multicolumn{2}{c}{Subject 9} & \multicolumn{2}{c}{Subject 10} & \multicolumn{2}{c}{Average} \\
			\cline{3-24} 
			& & top-1 & top-5 & top-1 & top-5 & top-1 & top-5 & top-1 & top-5 & top-1 & top-5 & top-1 & top-5 & top-1 & top-5 & top-1 & top-5 & top-1 & top-5 & top-1 & top-5 & top-1 & top-5 \\
			\hline 
			& BraVL\cite{du2023decoding}& 6.11& 17.89& 4.90& 14.87& 5.58& 17.38& 4.96& 15.11& 4.01& 13.39& 6.01& 18.18& 6.51& 20.35& 8.79& 23.68& 4.34& 13.98& 7.04& 19.71& 5.82& 17.45\\
			& NICE-GA\cite{song2023decoding}& 15.20& 40.10& 13.90& 40.10& 14.70& 42.70& 17.60& 48.90& 48.90& 29.70& 16.40& 44.40& 14.90& 43.10& 20.30& 52.10& 14.10& 39.70& 19.60& 46.70& 15.60& 42.80\\
			{200-way}
			& ATMS\cite{li2024visual}& 21.00& 51.50& 24.50& 54.00& 27.00& 61.00& 31.50& 60.50& 19.50& 44.50& 24.00& 59.50& 25.50& 57.00& 37.00& 72.00& 26.00& 53.50& 34.00& 69.50& 27.00& 58.50\\
			& MB2C\cite{wei2024mb2c}& 23.67& 56.33& 22.67& 50.50& 26.33& 60.17& 34.83& 65.00& 21.33& \textbf{53.00}& 31.00& \textbf{62.33}& 25.00& 54.83& 39.00& 69.33& 27.50& 59.33& 33.17& 70.83& 28.45& 60.17\\
			& \textbf{Neural-MCRL}& \textbf{27.50}& \textbf{64.00}& \textbf{28.50}& \textbf{61.50}& \textbf{37.00}&\textbf{69.00}& \textbf{35.00}& \textbf{66.00}& \textbf{22.50}& 51.50& \textbf{31.50}& 61.00& \textbf{31.50}& \textbf{62.50}& \textbf{42.00}& \textbf{74.50}& \textbf{30.50}& \textbf{59.50}& \textbf{37.50}& \textbf{71.00}& \textbf{32.25}& \textbf{64.15}           \\ 
			\hline
			& BraVL\cite{du2023decoding}& 14.80& 41.50& 12.88& 39.15& 15.0& 40.85& 12.35 & 36.45& 10.45& 33.77& 15.1& 41.17& 15.12& 42.38& 20.32& 49.83& 10.55&34.1& 16.75& 43.6& 14.33& 40.28\\
			50-way 
			& ATMS\cite{li2024visual}&38.50&76.50&34.50&\textbf{84.50}&43.50&80.00&52.50&81.50&32.00       &62.50&48.50&81.50&36.50 &72.50 & 65.50&86.50&55.50&82.50&57.00&85.50    &46.40&79.35\\
			& MB2C\cite{wei2024mb2c}&41.33 & 83.33 & 38.67 & 80.67 & 48.67 & 84.67 & \textbf{56.00} & 81.40 & 36.33 & 70.00 & \textbf{54.67} & 82.00 & 40.33 & 80.67 & 67.67 & 83.50 & 53.33 & 84.33 & 58.67 & 87.25 & 49.57 & 81.88\\
			& \textbf{Neural-MCRL}& \textbf{46.50}& \textbf{87.50}& \textbf{42.50}&84.00& \textbf{52.50}& \textbf{87.00}& 54.50& \textbf{82.50}& \textbf{38.50}& \textbf{72.50}& 54.00& \textbf{85.00}& \textbf{45.50}& \textbf{83.50}& \textbf{69.50}& \textbf{89.50}& \textbf{57.50}& \textbf{87.50}& \textbf{64.50}& \textbf{88.00}& \textbf{52.55}& \textbf{84.70}\\ 
			\hline\noalign{\hrule height 1.2pt}
			&  & \multicolumn{22}{c}{\normalsize Subject-independent: leave one subject out for test}\\
			\cline{3-24} 
			Type& Method & \multicolumn{2}{c}{Subject 1} & \multicolumn{2}{c}{Subject 2} & \multicolumn{2}{c}{Subject 3} & \multicolumn{2}{c}{Subject 4} & \multicolumn{2}{c}{Subject 5} & \multicolumn{2}{c}{Subject 6} & \multicolumn{2}{c}{Subject 7} & \multicolumn{2}{c}{Subject 8} & \multicolumn{2}{c}{Subject 9} & \multicolumn{2}{c}{Subject 10} & \multicolumn{2}{c}{Average} \\
			\cline{3-24} 
			&& top-1& top-5& top-1& top-5& top-1& top-5& top-1& top-5& top-1& top-5& top-1& top-5& top-1& top-5& top-1& top-5& top-1& top-5& top-1& top-5& top-1& top-5\\
			\hline
			& BraVL\cite{du2023decoding}& 2.30& 7.99& 1.49& 6.32& 1.39& 5.88& 1.73& 6.65& 1.54& 5.64& 1.76& 7.24& 2.14& 8.06& 2.19& 7.57& 1.55& 6.38& 2.30& 8.52& 1.84& 7.02\\
			& NICE-SA\cite{song2023decoding}& 7.00& 22.60& 6.60& 23.20& 7.50& 23.70& 5.40& 21.40& 6.40& 22.20& 7.50& 22.50& 3.80& 19.10& 8.50& 24.40& 7.40& 22.30& 9.80& 29.60& 7.00& 23.10\\
			200-way 
			&ATMS\cite{li2024visual}&9.50&24.50&11.50&\textbf{33.50}&8.5&29.50&11.50&30.00&8.50&24.00&10.50  &27.50 &8.00&26.50&13.50&30.50&9.50&27.50&12.50&37.00&9.55&29.05 \\
			& MB2C\cite{wei2024mb2c}& 10.50& 28.17& 11.33& 32.83& 8.83& 27.67& \textbf{13.67}& 33.50& 10.67& 27.50& 12.17& 33.17& 11.50& 31.83& 12.00& 32.17& 12.17& 31.33& 16.17& 37.17& 11.90& 32.03\\
			& \textbf{Neural-MCRL}& \textbf{13.00}& \textbf{31.50}& \textbf{12.00}&30.50& \textbf{14.50}& \textbf{35.50}& 12.50& \textbf{35.00}& \textbf{11.50}& \textbf{29.00}& \textbf{13.50}& \textbf{35.50}& \textbf{14.00}& \textbf{36.00}& \textbf{18.50}& \textbf{38.50}& \textbf{13.50}& \textbf{32.50}& \textbf{17.00}& \textbf{39.00}& \textbf{14.00}& \textbf{34.30}\\ 
			\hline
			&BraVL\cite{du2023decoding}&6.38&22.98&4.98&20.7&3.92&17.8&5.6&18.6&4.67&19.38&5.65&23.08&6.25&24.12&6.02&23.9&4.58&18.7&5.85&22.8&5.39&21.2\\
			50-way 
			& ATMS\cite{li2024visual}& 22.50&64.50&32.50&71.00&19.50&59.50&28.50&61.50&23.00&60.50&19.50&53.50&20.50&55.50&27.00&50.50&21.00&56.50&30.50&72.00&24.45&60.50\\
			& MB2C\cite{wei2024mb2c}& 25.33&68.00&34.67&74.00&18.00&59.33&\textbf{29.33}&\textbf{63.33}&22.00&59.33&20.00&54.67&22.67&59.33&26.67&48.67&23.33&\textbf{62.67}&31.33&77.33&25.33&62.67\\
			& \textbf{Neural-MCRL}& \textbf{27.50}& \textbf{69.50}& \textbf{38.00}&\textbf{76.50}& \textbf{26.50}& \textbf{65.50}& 28.00& 62.00& \textbf{24.50}& \textbf{61.50}& \textbf{23.50}& \textbf{58.50}& \textbf{24.50}& \textbf{63.00}& \textbf{27.50}& \textbf{52.50}& \textbf{24.50}& 61.50& \textbf{33.50}& \textbf{78.50}& \textbf{27.90}& \textbf{64.80}\\ 
			\hline\noalign{\hrule height 1.2pt}
		\end{tabular}
	}
\end{table*}

To achieve rich semantic representations in our framework, we aim to enhance semantics through two strategies: inter-modal semantic consistency and intra-modal semantic completion.
\subsubsection*{\textbf{Inter-modal Semantic Consistency}}
EEG signals inherently vary due to cognitive processes and temporal dynamics \cite{chen2024bridging}, while textual descriptions of visual scenes highlight different semantic aspects. These descriptions can be generated by a semantic generator leveraging the OPT-2.7b model \cite{li2023blip} to ensure diverse semantic representations. To address this semantic heterogeneity in cross-modal alignment, we propose a class-wise feature aggregation strategy inspired by prototype learning and neural representation consistency \cite{xu2022semi}. For each class $y_i \in \mathcal{Y}s$, we aggregate multiple text embeddings corresponding to the same visual scene:
\begin{align}
	\tilde{Z}_{X} = \frac{1}{K}\sum_{k=1}^{K}Z_{X,k},
\end{align}
where $K$ is the number of text descriptions per class, and $Z_{X,k} \in \mathbb{R}^{F}$ represents the CLIP text embedding of the $k$-th description.
This aggregation yields prototype semantic representations that capture conceptual centroids while reducing description-specific noise. The resultant $\tilde{Z}_{X}$ serves as semantic anchors for $Z_E$ alignment, mimicking the brain's formation of stable concepts through repeated exposure.

\subsubsection*{\textbf{Intra-modal Semantic Completion}}
EEG signals contain complex semantic information, but projecting into shared embedding spaces may result in information degradation. This degradation is exacerbated by the inherently low SNR of EEG recordings and the stochastic nature of neuronal responses \cite{winterer1999cortical}. To preserve semantic integrity, we implement noise-based augmentation inspired by manifold learning and local intrinsic dimensionality \cite{amsaleg2015estimating}. Specifically, during training, EEG embeddings $\mathbf{Z}_E \in \mathbb{R}^d$ are perturbed with Gaussian noise:
\begin{align}
\tilde{Z}_E = Z_E + \theta, \quad \theta \sim \mathcal{N}(0, \sigma^2),
\end{align}
where $\sigma$ controls noise magnitude. This approach, grounded in noise-contrastive estimation, helps recover the underlying manifold structure of neural activity patterns \cite{churchland2012neural}. By employing L2-normalized embeddings on the unit hypersphere, Gaussian noise transforms discrete points into local spherical regions, enriching semantic representations. This aligns with neighborhood-preserving embedding theory \cite{he2005neighborhood}, where local structures encode semantic relationships. As a form of vicinal risk minimization, this augmentation ensures robust representations while maintaining semantic consistency for zero-shot recognition.

\subsection{EEG-Image-Text Representation Alignment (EITRA)}
EITRA addresses the semantic gap between different modalities by introducing modality-specific semantic interaction matrices. These matrices facilitate cross-modal interactions through semantic-guided attention mechanisms while preserving the unique characteristics of each modality.

\subsubsection*{\textbf{Semantic-guided Attention}}
The semantic connections between modalities are established through semantic-guided attention mechanisms. For the features from EEG and images, we introduce learnable semantic interaction matrices $M_{E}$ and $M_{I}$ that actively query the modality relationships:
\begin{equation}
	Q_E = M_E W_Q, \quad Q_I = M_I W_Q,
\end{equation}
\begin{equation}
	K_{X} = \tilde{Z}_{X}W_K, \quad V_{X} = \tilde{Z}_{X}W_V,
\end{equation}
where $W_Q$, $W_K$, and $W_V$ are learnable projection matrices. The semantic interactions are then computed as:
\begin{equation}
	\mathbf{A}_E = \text{softmax}(\frac{Q_E K_{X}^T}{\sqrt{d}})V_{X}, \quad \mathbf{A}_I = \text{softmax}(\frac{Q_I K_{X}^T}{\sqrt{d}})V_{X},
\end{equation}
where $\mathbf{A}_E$ and $\mathbf{A}_I$ represent the semantically enriched EEG and image features, respectively, both guided by textual knowledge. This semantic-guided attention mechanism ensures that both modalities are anchored to the same semantic space through textual knowledge.

\subsubsection*{\textbf{Cross-modal Semantic Alignment}}
The semantically enriched representations are integrated through cross-modal semantic alignment. We introduce a cross-modal interaction matrix $\mathbf{M}_C \in \mathbb{R}^{M \times d}$ to facilitate EEG-Image alignment:
\begin{equation}
	Q_C = \mathbf{M}_C W_Q, \quad K_A = [\mathbf{A}_E; \mathbf{A}_I] W_K, \quad V_A = [\mathbf{A}_E; \mathbf{A}_I] W_V,
\end{equation}
\begin{equation}
	Z_{E} = \text{\text{ResNet}}(\text{softmax}(\frac{Q_C K_A^T}{\sqrt{d}})V_A),
\end{equation}
where $[;]$ denotes concatenation. The resulting architecture creates a hierarchical semantic bridge through learnable interaction matrices that actively query and align modality-specific features. By leveraging text as an intermediate semantic anchor and employing dedicated interaction matrices, our framework captures both fine-grained neural patterns and high-level visual concepts while maintaining modality-specific characteristics. The loss
function for our model is expressed as:
\begin{equation}
	\mathcal{L}_{con}(z_1, z_2) = -\frac{1}{N}\sum_{i=1}^N \left[
	\log\frac{\exp(z_1^i \cdot z_2^i/\tau)}{\sum_{j=1}^N \exp(z_1^i \cdot z_2^j/\tau)} 
	\right],
\end{equation}
\begin{equation}
	\mathcal{L}_{oss} = \eta\mathcal{L}_{con}(Z_{E}, Z_I) + (1-\eta)\mathcal{L}_{con}(Z_{E}, Z_X).
\end{equation}
where $\tau$ is the temperature parameter, $\eta$ controls the balance between image and text alignment.

\section{Experiments and Results}

\subsection{Experimental Details}
Our experiments used the THINGS-EEG dataset \cite{gifford2022large}, which includes EEG recordings from 10 participants via a 63-channel system with Rapid Serial Visual Presentation. The dataset comprises 1,654 training categories (10 exemplars per category, four presentations per exemplar) and 200 testing categories (one exemplar per category, 80 presentations per exemplar). The raw 1000Hz EEG acquisitions underwent preprocessing steps, including temporal segmentation, baseline normalization using 200ms pre-stimulus intervals, and frequency downsampling to 250Hz.

The experiments ran on a single NVIDIA RTX 3090 GPU. We trained the models at the across-subject and within-subject levels using the Adam optimizer \cite{kingma2014adam}, with an initial learning rate of \(3 \times 10^{-4}\) and batch sizes of 128. To ensure fairness, all models were configured with identical hyperparameters. Performance was evaluated on the zero-shot test dataset at the end of each training epoch during the training process.

\subsection{Overall Performance and Ablation Studies}
\subsubsection*{\textbf{\textit{N}-way Zero-shot Classification Performance}}
Our experimental evaluation covered zero-shot classification in both subject-dependent and subject-independent scenarios, as outlined in Table \ref{acc}. Neural-MCRL achieved a top-1 accuracy of 52.55\% and a top-5 accuracy of 84.70\% in 50-way subject-dependent classification while maintaining a top-1 accuracy of 32.25\% and a top-5 accuracy of 64.15\% in 200-way scenarios. In subject-independent evaluations, despite challenges posed by cross-subject variability, Neural-MCRL achieved a top-1 accuracy of 27.80\% and a top-5 accuracy of 64.80\% for 50-way classification, and a top-1 accuracy of 14.00\% and a top-5 accuracy of 34.30\% in 200-way classification. Comparative analysis against contemporary approaches, including BraVL \cite{du2023decoding}, NICE variants \cite{song2023decoding}, ATMS \cite{li2024visual} and MB2C \cite{wei2024mb2c}, showcased the superior performance of Neural-MCRL, setting new benchmarks in neural decoding and neural-semantic understanding.

\begin{table}[!t]
    \caption{Ablation Studies with regard to Components in Neural-MCRL.}
    \label{table_ablation}
    \centering
    \setlength{\tabcolsep}{2.5pt}
    \renewcommand{\arraystretch}{0.9}
    \begin{tabular}{@{}c|c|cc@{}}
        \toprule
        \multicolumn{2}{c|}{\textbf{Component Settings}} & \textbf{\makecell{Subject\\dependent\\\textit{(Top1-/Top-5)}}} & \textbf{\makecell{Subject\\independent\\\textit{(Top1-/Top-5)}}} \\
        \midrule
        \multirow{2}{*}{\makecell[c]{NESTA\\Components}} & w/o Subject-Specific & 28.10$/$60.55\% & 12.05$/$31.75\% \\
        & w/o Neural-Spectral & 25.55$/$55.85\% & 10.05$/$28.95\% \\
        \midrule
        \multirow{2}{*}{Semantic Enhancement} & w/o Consistency & 30.15$/$60.45\% & 12.75$/$32.25\% \\
        & w/o Completion & 30.35$/$61.35\% & 13.00$/$32.75\% \\
        \midrule
        EITRA & w/o Alignment & 26.05$/$57.75\% & 10.35$/$29.15\% \\
        \midrule
        \multicolumn{2}{c|}{\textbf{Neural-MCRL}} & \textbf{32.25$/$64.15\%} & \textbf{14.00$/$34.30\%}\\
        \bottomrule
    \end{tabular}
\end{table}

\subsubsection*{\textbf{NESTA Component-wise Performance Evaluation}}
As shown in Table \ref{table_ablation}, the ablation results validated the designed component's rationale in NESTA. The Subject-Specific Layer effectively captures individual neural patterns and spatial channel interactions, while Neural-Spectral Adaption Block confirms the importance of adaptive spectral processing. 

\begin{figure}[t]
	\centering
	\includegraphics[width = 0.38\textwidth]{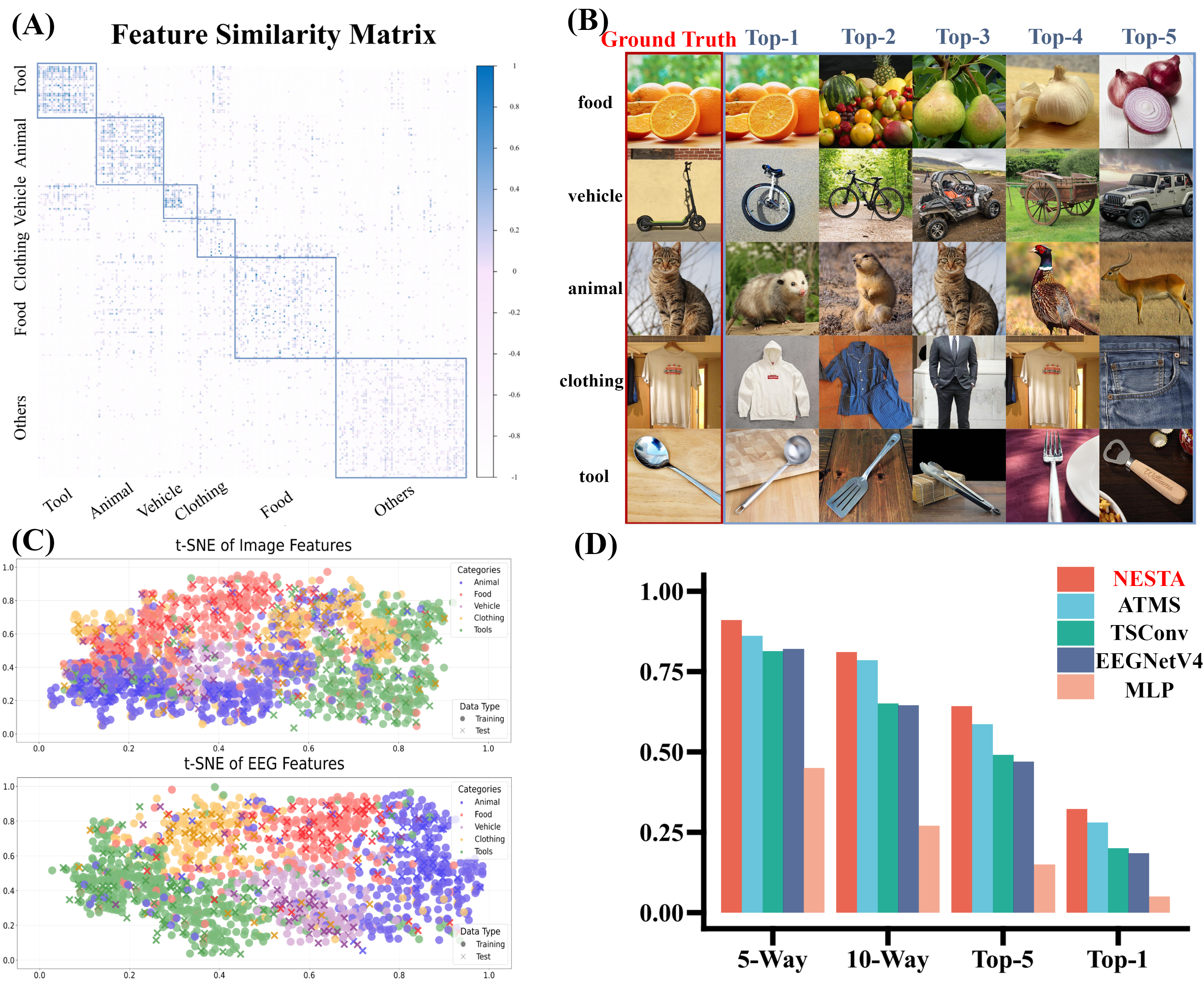}
	\caption{Semantic similarity analysis and visualization. (A) Cosine similarity of 200 concepts in the test set. The results calculated by the trained models of 10 subjects were averaged, and all the concepts were rearranged into six categories. (B) Comparison between ground truth (first column) and the top-5 predicted. (C) Visualization in the latent space of EEG/image by t-SNE. (D) Average in-subject accuracy across different EEG Encoders.}
	\label{sim}
\end{figure}

\subsubsection*{\textbf{Performance Evaluation of Semantic Enhancement and EITRA}}
Table \ref{table_ablation} revealed that the EITRA alignment module is crucial for establishing robust cross-modal bridges, while the semantic enhancement module provides complementary benefits in preserving semantic information. This validates our hypothesis that semantic enhancement and alignment strengthen the model's capability in capturing semantic information across modalities.

\subsection{Semantic Similarity and Visualization Analysis}
To evaluate the semantic comprehension capability of Neural-MCRL, we conducted representational similarity analysis by categorizing the 200 test concepts into six categories: animal, food, vehicle, tool, clothing, and others. The cosine similarity matrix reveals distinct intra-category aggregation, indicating strong semantic alignment between EEG and visual representations in Figure \ref{sim} (A). Visual inspection of the top-5 predictions, as shown in Figure \ref{sim} (B), demonstrates semantic consistency, where predicted items share conceptual relationships with the ground truth, confirming that our model captures meaningful semantic features rather than just low-level visual properties. As shown in Figure \ref{sim} (C), visualization in the latent space of EEG/image by t-SNE \cite{van2008visualizing}.

\subsection{Performance Comparison across EEG Encoders}
In this section, we compared  the 5-way, 10-way, Top-1 and Top-5 accuracy of 200-way across different EEG embedding encoders, such as our NESTA, ATMS \cite{li2024visual}, TSConv \cite{song2023decoding}, EEGNetV4 \cite{lawhern2018eegnet} and MLP. As shown in Figure \ref{sim} (D), the proposed NESTA encoder demonstrates remarkable superiority over the state-of-the-art EEG encoders across all evaluation scenarios.

\section{Conclusion}
In this paper, we introduced Neural-MCRL, a novel framework for EEG-based visual decoding that advances beyond simple strategies in multimodal contrastive learning. Neural-MCRL has three key components: the specialized EEG encoder NESTA, designed to capture adaptive spectral and subject-specific representations; a semantic enhancement module that improves both intra-modal semantic completeness and cross-modal consistency; and the EITRA module, which enhances semantic alignment among EEG, image and text representations. Our experiments demonstrate substantial gains in zero-shot classification accuracy and improved generalization across subjects. By advancing neural visual decoding beyond fragmentary mappings to deeply integrated, contextually aware representations, Neural-MCRL lays a solid foundation for future research in brain-computer interfaces and interpretable neural semantic understanding.

\bibliographystyle{IEEEbib}
\bibliography{icme2025_template_anonymized.bbl}

\vspace{12pt}

\end{document}